%% file: POLIS-Bench-main.tex
\documentclass[final,3p,times]{elsarticle}

\usepackage{amssymb}
\usepackage{amsmath}
\usepackage{colortbl,xcolor}
 \usepackage{booktabs}

\usepackage{xcolor}

\journal{Knowledge-Based Systems}
\usepackage[utf8]{inputenc}
\DeclareUnicodeCharacter{FF1A}{:}
\usepackage[skip=8pt plus 2pt minus 1pt]{parskip}
\usepackage[hyphens]{url}
\usepackage[hidelinks]{hyperref}
\makeatletter
\renewcommand\@biblabel[1]{[#1]}
\makeatother
\usepackage{booktabs,tabularx,array,colortbl,xcolor,adjustbox}
\newcolumntype{C}{>{\centering\arraybackslash}X}

\definecolor{cool1}{RGB}{239,243,255}
\definecolor{cool2}{RGB}{198,219,239}
\definecolor{cool3}{RGB}{158,202,225}
\definecolor{cool4}{RGB}{107,174,214}
\definecolor{cool5}{RGB}{ 33,113,181}
\newcommand{\coldcell}[1]{%
  \begingroup\def\val{#1}%
  \ifdim \val pt < 0.4pt \cellcolor{cool1}%
  \else\ifdim \val pt < 0.55pt \cellcolor{cool2}%
  \else\ifdim \val pt < 0.65pt \cellcolor{cool3}%
  \else\ifdim \val pt < 0.75pt \cellcolor{cool4}%
  \else \cellcolor{cool5}\fi\fi\fi\fi
  #1\endgroup}
\usepackage{multirow}
\usepackage[table,xcdraw]{xcolor}
\definecolor{bestgreen}{HTML}{E5B5B5}   
\definecolor{secondgreen}{HTML}{AFC3D8}

\usepackage{graphicx} 
\usepackage{float} 
\usepackage{subfigure} 
\usepackage[skip=6pt]{caption}  
\usepackage{wrapfig}
\usepackage{cleveref}

\begin{document}

\begin{frontmatter}




\title{\textbf{POLIS-Bench}: Towards Multi-Dimensional Evaluation of LLMs for Bilingual Policy Tasks in Governmental Scenarios}

\cortext[cor1]{Corresponding author}
\author[add1]{Tingyue Yang}
\ead{2023170201018@std.uestc.edu.cn}
\author[add1]{Junchi Yao}
\ead{2022160101012@std.uestc.cn}
\author[add1]{Yuhui Guo\corref{cor1}}
\ead{gyh@uestc.edu.cn}
\author[add1]{Chang Liu}
\affiliation[add1]{organization={University of Electronic Science and Technology of China}
}


\begin{abstract}
 We introduce \textsc{POLIS-Bench}, the first rigorous, systematic evaluation suite designed for LLMs operating in governmental bilingual policy scenarios. Compared to existing benchmarks, \textsc{POLIS-Bench} introduces three major advancements. (i)\textbf{Up-to-date Bilingual Corpus:} We construct an extensive, up-to-date policy corpus that significantly scales the effective assessment sample size, ensuring relevance to current governance practice. (ii) \textbf{Scenario-Grounded Task Design:} We distill three specialized, scenario-grounded tasks---Clause Retrieval \& Interpretation, Solution Generation, and the Compliance Judgment---to comprehensively probe model understanding and application. (iii)\textbf{Dual-Metric Evaluation Framework:} We establish a novel dual-metric evaluation framework combining semantic similarity with accuracy rate to precisely measure both content alignment and task requirement adherence. A large-scale evaluation of over 10 state-of-the-art LLMs on \textsc{POLIS-Bench} reveals a clear performance hierarchy where reasoning models maintain superior cross-task stability and accuracy, highlighting the difficulty of compliance tasks. Furthermore, leveraging our benchmark, we successfully fine-tune a lightweight open-source model. The resulting POLIS series models achieves parity with, or surpasses, strong proprietary baselines  on multiple policy subtasks at a significantly reduced cost, providing a cost-effective and compliant path for robust real-world governmental deployment.
\end{abstract}


\begin{highlights}
\item \textbf{Systematic Bilingual Benchmarking:} Introduction of \textsc{POLIS-Bench}, the first systematic, rigorous evaluation suite for LLMs in governmental bilingual (Chinese/English) policy scenarios, designed to address issues of "false success" and compliance violation in policy generation.
\item \textbf{Scenario-Grounded Task Distillation:} Distillation of three specialized, scenario-based tasks (Clause Retrieval \& Interpretation, Solution Generation, and Compliance Judgment) that comprehensively probe models' policy understanding and adherence to compliance constraints.
\item \textbf{Dual-Metric Evaluation Framework:} Establishment of a novel dual-metric evaluation framework combining semantic similarity and LLM.Judge accuracy to quantitatively assess both lexical alignment and genuine task correctness in policy contexts.
\item \textbf{Cost-Efficient Task Adaptation:} Successful task-aligned fine-tuning of lightweight POLIS series open-source models, achieving parity with or surpassing strong proprietary baselines (e.g., GPT-4 series) on key policy subtasks while maintaining general capabilities and significantly lowering deployment costs.
\end{highlights}

\begin{keyword}
Large Language Models \sep  Natural Language Processing \sep Evaluation Benchmark \sep  Public Governance


\end{keyword}

\end{frontmatter}


 \begin{figure}[h] 
    \centering 
    \includegraphics[width=0.63\textwidth]{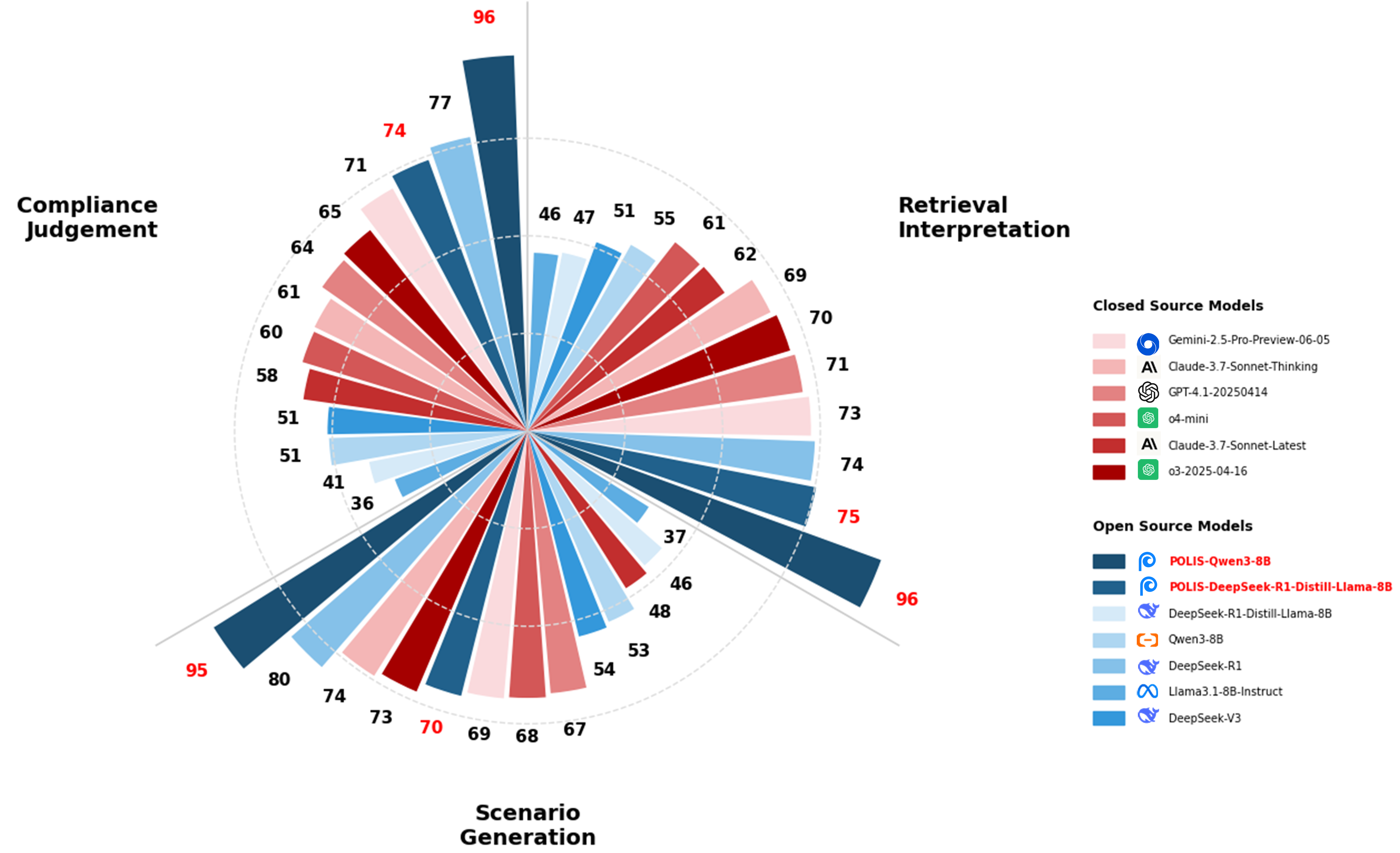} 
    \caption{Comparison of accuracy rate across 15 state-of-the-art models and POLIS series models.}
    \label{Pipeline} 
 \end{figure}

\input{Section/introduction}

\input{Section/relatedwork}
\input{Section/method}

\input{Section/experiment}
\input{Section/conclusion}
\bibliographystyle{elsarticle-num-names} 
 \bibliography{POLIS-Bench-main}

\appendix
\input{Section/appendix}

\end{document}

%% file: Section/introduction.tex
\section{Introduction}
In recent years, large language models (LLMs) have been widely adopted for their strengths in text modeling, information retrieval, and related abilities in e-Commerce \cite{1,XU2002485}, healthcare \cite{qiu2024llm,RAMOSSOTO2025114468}, and education \cite{chu2025llm,DUAN2024112118}. Meanwhile, with the rise of digital government, LLMs have attracted growing attention in public governance. A survey of 142 U.S. government departments reported that nearly 45 percent planned to deploy or had already deployed AI algorithms in governance tasks \cite{Engstrom_Ho_Sharkey_Cuéllar_2020}. However, while LLMs offer new technical leverage, their risks cannot be ignored: repetition \cite{Yao_2025}, bias, and data contamination \cite{liu2023trustworthy}, as well as potential public safety and ethical concerns \cite{Li_Liu_Gao_Buntine_2023,8862835}. These issues can distort understanding during task execution and lead to negative social impacts.


However, current LLMs often overlook key properties of policy--oriented generation. Compared with ordinary text, policy documents are typically lengthy, structurally complex, and dense with specialized terminology, requiring precise retrieval and interpretation across articles and contexts. In addition, policy analysis must strictly adhere to policy goals and compliance constraints. Ignoring these requirements frequently may lead to ``false success'', where an answer appears reasonable on the surface yet departs from the intent of the provisions or overlooks compliance boundaries.
 

Current benchmark studies specific to public--sector governance remain scarce, and existing evaluations of models on governance--oriented generation tasks often lack sound task design and evaluation methods \cite{liu2025evaluation}. Concretely, (i) the effective sample sizes for generation tasks are small, undermining representativeness and stability; (ii)  task categories are too coarse to cover the diverse real world policy scenarios; and (iii) evaluation methods still rely primarily on manual sampling and scoring, which limits scale and authority.

To address these gaps, we introduce \textbf{\textsc{POLIS-Bench}}---\textbf{P}olicy--\textbf{O}riented \textbf{L}anguage \textbf{I}ntelligence \textbf{S}uite \textbf{Bench}mark, a rigorously selected language intelligence suite for policy domains that standardizes corpus, tasks, and dual--metrics evaluation into one benchmark and with three key improvements: \textbf{First}, to address the problem of small actual assessment sample size of existing policy evaluation benchmarks, we build an up--to--date and realistic policy dataset aggregating recent Chinese--English policy texts from diverse sources and continually expanding the corpus to reflect current governance practice.\textbf{ Second}, to address the problem of rough classification of existing policy evaluation benchmark tasks, we designed scenario--based tasks. For each policy, we distill three task types---clause retrieval and interpretation, solution generation for concrete problems, and compliance judgment. \textbf{Third}, to address the problem of single and inefficient evaluation of existing policy evaluation benchmarks, we propose a dual--metric evaluation framework that combines semantic similarity with LLMJudge accuracy: cosine similarity measures the lexical closeness between model outputs and references, while LLMJudge assesses, in task context, whether an output satisfies the problem requirements.


In a large--scale evaluation of more than 10 state--of--the--art LLMs, we observe a clear performance hierarchy. \textbf{Firstly}, the closed--source reasoning models steadily dominate on multiple tasks and maintain higher consistency in cross--language and cross--task scenarios which demonstrates strong integrated reasoning capabilities. In contrast, though closed--source chat models have good language fluency and clause retrieval, their correctness and robustness usually lag behind reasoning--based routes in complex scenarios. \textbf{Secondly}, the open--source reasoning models perform relatively well, especially in the Chinese side and scenario--based tasks, but their overall stability is still far from that of the closed--source strong baseline model. Comparatively, open--source chat models are more prone to bias in cross--clause evidence integration and compliance boundary judgment, and get a lower score overall. Based on the evaluation results, we selected two high--performing open--source baseline models for joint multi--task \textit{LoRA} fine--tuning. Through a comprehensive analysis of the three types of tasks and two metrics, we found that the fine--tuned models can significantly improve semantic alignment and cross--clause localization capabilities in clause retrieval and interpretation. In addition, the fine--tuned model reduces the number of responses that are irrelevant to policy objectives or too general, and reduces unnecessary redundancy in scenario generation. Meanwhile, the fine--tuned model can identify constraint boundaries more robustly in compliance judgment and significantly reduce the incidence of non--compliant proposals. Overall, the fine--tuned model provides more accurate and relevant answers to questions in the government generation task, and matches or exceeds the strong closed--source baseline (e.g., GPT--4 series) in several subtasks, while maintaining smaller parameter sizes and lower calling costs, which significantly improves the deployment efficiency and feasibility of the policy task. Our contributions are:

 \begin{figure}[t] 
    \centering 
    \includegraphics[width=1\textwidth]{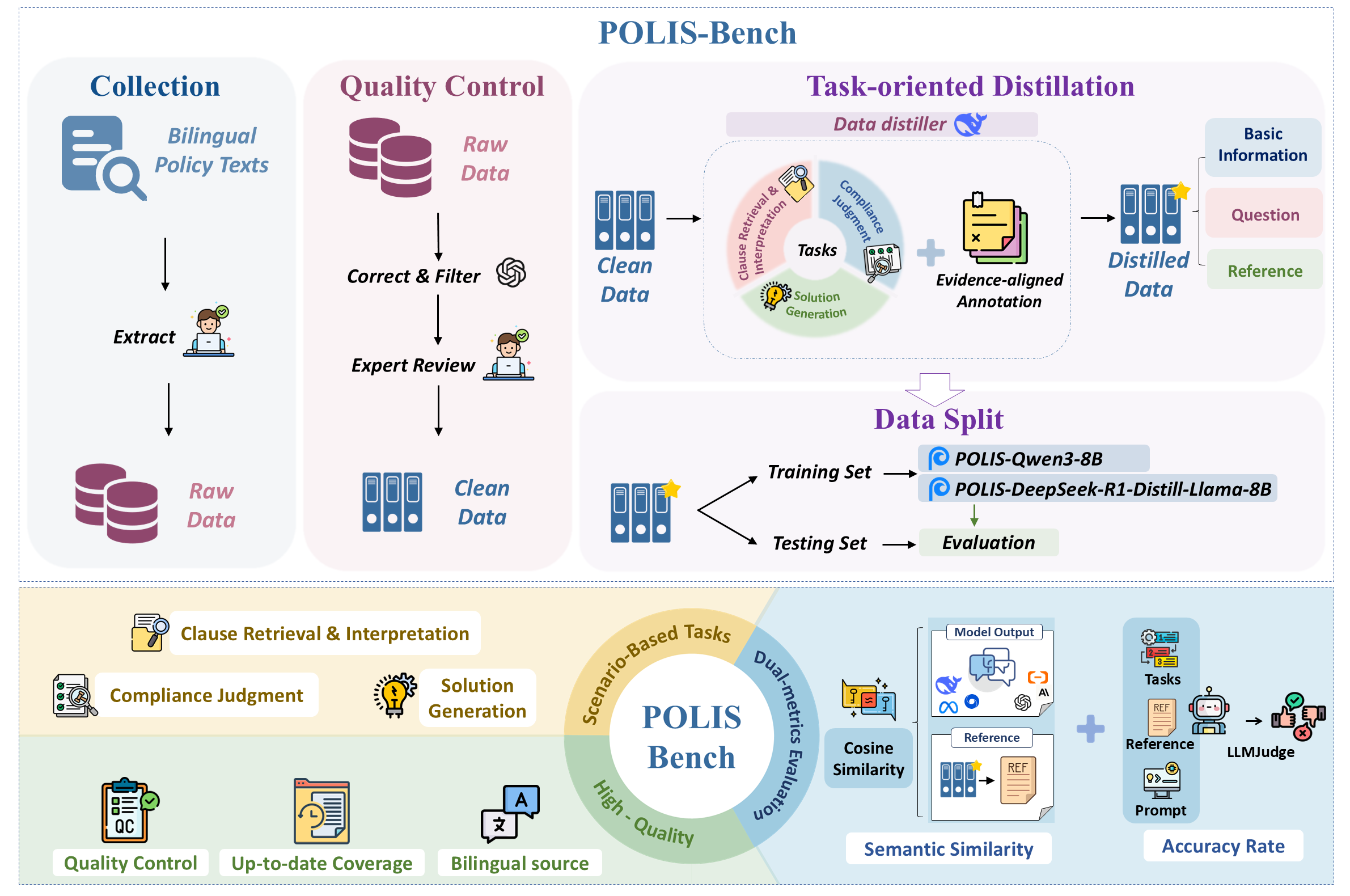} 
    \caption{\textbf{Overview of our POLIS-Bench.} (i) The diagram \textbf{above} illustrates construction pipeline of \textsc{POLIS--Bench}. (ii) The diagram \textbf{below} illustrates the three key characteristics of the \textsc{POLIS--Bench}.}
    \label{Pipeline} 
 \end{figure}
\begin{enumerate}
    \item \textsc{\textbf{POLIS--Bench.}} We present a systematic benchmark for policy--oriented generation that compiles up--to--date bilingual (Chinese/English) policies, distills each policy into three tasks---clause retrieval and interpretation, solution generation, and compliance judgment---and offers a unified, reproducible evaluation pipeline.
    \item \textbf{Dual--metrics evaluation framework} We establish an evaluation system with two metrics, namely semantic proximity and task correctness. The first measures how closely a model’s output aligns with the reference answer in content and expression. The second assesses, within concrete task contexts, whether the answer satisfies the requirements of the problem and adheres to the compliance constraints. The system is quantitatively clear and can be directly migrated and extended in bilingual and multi--tasking scenarios.
    \item \textbf{Lightweight model adaptation} Using \textbf{\textsc{POLIS-Bench}}, we successfully fine--tune a lightweight open--source model. The resulting POLIS--Qwen--8B delivers clear gains on all three tasks, achieves performance equal to or better than strong closed-source baselines at lower cost, and offers a cost--efficient, compliance--friendly path for real--world governmental deployment.
\end{enumerate}

%% file: Section/relatedwork.tex
\section{Related Work}
\textbf{The Emergence and Challenges of LLMs in Public Policy.}  In recent years, the application of LLMs within public policy has seen a marked increase in academic research, revealing the multifaceted value of these models throughout the entire policy life cycle. Current studies suggest that LLMs are improving evidence--based decision--making and governance by enhancing data analysis and predictive capabilities, thereby contributing to the creation of public value \cite{charles2022artificial,Yaosocial_2025}. For instance, in the critical domain of public safety, researchers have applied neural network models \cite{wang2020csan} to dynamically forecast spatiotemporal crime data, assisting in the optimization of law enforcement resource allocation and risk regulation. Moreover, LLMs are considered a key driver in advancing e--Government modernization, improving service quality and operational efficiency, and achieving higher citizen acceptance in certain contexts compared to traditional human--delivered services \cite{ivic2022challenges,gesk2022artificial}. Additionally, natural language processing (NLP) technologies embedded within LLMs are transforming the policy analysis process itself, facilitating public participation through the automatic evaluation of large volumes of citizen feedback   \cite{romberg2024making} and serving as a powerful tool for policy analysis 
 \cite{safaei2024end} in trend identification and information synthesis. More notably, LLMs have been employed in designing complex economic policies. The ``AI Economist'' framework \cite{zheng2022ai} uses deep reinforcement learning to design tax policies, achieving social welfare outcomes in simulations that surpass traditional economic models \cite{zheng2022ai}. Collectively, these studies demonstrate that LLMs have evolved from mere auxiliary tools to comprehensive transformative engines that optimize governance, enhance democratic interaction, and foster innovative policy solutions.

\textbf{From Universal to Specialized Benchmarks: The Unique Demands of Policy Analysis.} The evaluation of LLMs has evolved from universal benchmarks---such as multi--task challenges like GLUE and SuperGLUE \cite{chang2024survey,li2025application} and measures of general world knowledge like MMLU and its enhanced versions \cite{wang2024mmlu}—to more specialized domains. The scope of these universal benchmarks has expanded to include dimensions such as cross--lingual competence with XTREME \cite{bugliarello2022iglue,chen2024multi}, multimodal comprehension with IGLUE \cite{bugliarello2022iglue}, as well as evaluations of the "human--likeness" of generated text with TURINGBENCH \cite{uchendu2021turingbench} and personalized long--form content generation with LongLaMP \cite{kumar2024longlamp}, thereby comprehensively assessing the fundamental capabilities of models. However, the limitations of universal benchmarks have spurred the development of specialized evaluation frameworks for specific high-risk domains. In the legal domain, LexEval and LawBench \cite{li2024lexeval,fei2023lawbench} in the Chinese context have established multi--dimensional legal cognition frameworks, while in the English--speaking world, LegalBench and CaseHOLD \cite{guha2023legalbench,zheng2021does} collaborate with legal experts to focus on specific tasks such as case holding identification. In the financial sector, evaluations have evolved from early systems like FLUE \cite{shah2022flue} and PIXIU \cite{xie2023pixiu} to comprehensive frameworks such as FinBen \cite{xie2024finben}, which covers tasks from sentiment analysis to quantitative trading, and FinRL--Meta \cite{liu2022finrl}, which focuses on reinforcement learning--driven trading strategies. In the healthcare domain, FLamby \cite{ogier2022flamby} concentrates on privacy concerns within federated learning environments, while XLINGHEALTH \cite{jin2024better} is dedicated to addressing information disparities in cross--lingual medical question-answering. While these specialized benchmarks provide profound insights within their respective fields, they fundamentally differ from the core requirements of policy analysis. Evaluations in the legal, financial, and healthcare domains are largely centered on established rules, objective data, or scientific consensus, with a primary focus on factual accuracy and predictive precision. In contrast, the core task of policy analysis involves not only fact extraction but also the comprehension and articulation of competing ideological perspectives, the weighing of trade--offs, conducting social impact assessments, and providing policy recommendations. This unique demand for reasoning that incorporates argumentation, normative considerations, and value plurality is not adequately addressed by existing benchmarks, making the development of a dedicated policy analysis benchmark both necessary and urgent.

\textbf{Opportunities and Challenges in Government Affairs Evaluation.}    Compared to domains such as law, finance, and healthcare, the evaluation of LLMs within the government affairs domain is still in its early stages. Pioneering efforts, such as those presented by Liu in MSGABench \cite{liu2025evaluation}, have made significant contributions by constructing a comprehensive evaluation framework that highlights challenges related to model reliability, security, and other factors, providing a critical foundation for future research. However, while these efforts have effectively diagnosed macro--level deficiencies in model capabilities, significant limitations remain in fine--grained, in--depth evaluations of policy--related text generation tasks, particularly in task design and evaluation methodology.

Building upon these research experiences and acknowledging existing gaps, this paper proposes \textbf{\textsc{POLIS--Bench}}, a benchmark inspired by the development of evaluation frameworks in high--risk domains such as healthcare \cite{hou2025msdiagnosis} and responding to the call for multi--criteria decision-based evaluations \cite{gjorgjevikj2025user}. \textbf{POLIS--Bench} aims to complement existing government affairs benchmarks by addressing data timeliness and bilingual coverage, while also representing a critical step in refining task design and evaluation methodologies. The goal is to provide a more accurate and responsible framework for assessing the application of LLMs in the field of public policy, thereby supporting the responsible and effective use of these models.

%% file: Section/method.tex
\section{Method}
\subsection{\textbf{Overview}}
  We introduce \textsc{POLIS--Bench}, the first benchmark for policy analysis tasks, designed to evaluate LLMs’ reading, reasoning, and compliance aware generation in real governmental policy contexts. Its contributions are threefold: (i) a comprehensive, up--to--date bilingual (Chinese/English) policy corpus that tracks current governance practice; (ii) three scenario--grounded task types that jointly probe retrieval, generation, and compliance; and (iii) a dual-metric evaluation framework comprising two metrics---semantic proximity and task correctness---and providing a unified, reusable pipeline. Building on this benchmark, we perform multi--task joint fine--tuning on open--source backbones; the fine--tuned models match or surpass strong closed--source baselines across multiple subtasks while delivering lower cost and stronger deployability.

\textbf{Dataset Perspective}  \textsc{POLIS--Bench} compiles bilingual policy texts from approximately the past five years, sourced from officially released policy documents, and is continuously expanded to reflect current governance practice. For each policy, we design three tasks---Clause Retrieval \& Interpretation, Solution Generation, and Compliance Judgment--- and perform data distillation to obtain 1,000 items for each of the three categories. All questions are generated from the corresponding policy texts, and the answer--evidence spans in the texts are precisely annotated and used as supervision for training and evaluation. 

\textbf{Evaluation}  Within \textsc{POLIS--Bench}, we design a dual--metrics evaluation framework: (i) \textbf{Semantic similarity} uses cosine similarity to measure the degree of similarity between model outputs and reference answers; (ii) \textbf{Accuracy rate} uses LLMJudge to determine correctness. Compared with relying solely on cosine similarity as a single machine score, we introduce LLMJudge scoring to provide a comprehensive evaluation of core semantic coverage, the legitimacy of explanations, and information veracity, thus more faithfully reflecting overall performance on policy tasks in governmental scenarios.



\subsection{\textbf{Data Construction}}
\textbf{Up--to--date Data Collection} We systematically collect long--form bilingual policy texts released over the past five years from official government websites, reflecting current governance practice and keeping the corpus continuously expanded. Considering that some originals are lengthy and exceed the input budget, we apply length--constrained, standardized segmentation to the original texts, ultimately obtaining 500 normalized long--text samples in Chinese and 500 in English (1,000 in total) as the raw database.

\textbf{Quality Control} To ensure usability and consistency, we apply structured quality control to the original result. First, we use LLMs to correct and filter the raw dataset: it validates and parses JSON fields, removes samples that are unparseable or missing key fields, and deduplicates when duplicates or templated outputs occur yielding samples that are clear and structurally uniform. Then, we conducted the manual sampling inspection to review and correct the machine-screened samples. The resulting Clean Data serve as the input to the subsequent task-oriented distillation.

\textbf{Task--oriented Distillation} To convert general policy texts into evaluable task samples, we perform instruction--driven data distillation: given each policy’s title and body, we use a prompt template to generate, in order, three question types corresponding to the three tasks--Clause Retrieval \& Interpretation, Solution Generation, and Compliance Judgment--and require the precise textual source for each question (i.e., the concrete phrasing from the original policy). This directly supports subsequent capability evaluation for different tasks: clause--type questions align with policy provisions and key interpretive points; solution--type questions align with policy goals and execution pathways; compliance--type questions align with boundary and rule requirements.

\textbf{Evidence aligned Annotation} During distillation we annotate, for each question, the answer--evidence spans drawn from the long--form policy text as the alignment signal for training and evaluation: Clause Retrieval \& Interpretation aligns to the relevant provisions and their key interpretive statements; Solution Generation aligns to policy segments related to solution points; Compliance Judgment aligns to clause statements that delimit constraint boundaries. This annotation scheme preserves traceability between samples and originals, enabling evaluation to focus on the core goal of ``answering on the basis of the policy text.''



\textbf{Construction Pipeline} Data construction follows a unified pipeline to ensure compliant sources, consistent structure, and usable formats. First, obtain policy titles and bodies and standardize them into long text inputs. Second, for overly long originals, conduct sentence level, length constrained splitting and fragment processing to ensure cross language consistency and comparability. Then, organize task samples in a ``question--answer evidence'' structure, so that each sample can be located in the original policy text. The entire process focuses on mapping original policy texts to task-specific samples, ensuring that all task instances are derived from the same underlying corpus and remain auditable and reusable.


\textbf{Evaluation Readiness \& Data Split} The final data is released as standardized sample units: each sample contains policy metadata, task type, the question text, and its corresponding source annotation. The three tasks share the same construction and annotation procedure and connect directly to our evaluation framework. Meanwhile, from the full set of 3,058 instances, we randomly sampled 2,500 instances as the training set for LoRA fine-tuning and 558 instances as the test set for evaluation only. Through this construction pipeline and annotation scheme, the dataset attains a consistent standard in coverage, structure, and evaluation readiness, providing a solid foundation for comparing different models and settings.

\subsection{\textbf{Evaluation}}

We used two complementary types of metrics to evaluate the state--of--the--art models' performance in
governmental scenarios: \textbf{semantic similarity} to measure semantic proximity, and \textbf{accuracy rate} to judge the correctness of the model responses in terms of the dimensions of the task requirements.

\textbf{Semantic similarity}  In the semantic similarity setting, we use cosine similarity between the vector representations of the model output and the reference answer to
quantify semantic closeness. To accommodate long texts, we first perform
sentence level segmentation on both the model output and the reference,
and concatenate the resulting sentences into several segments,
which are then compared in pairs. For any sentence pair in the vector space, let the vectors be $\mathbf{v}_1$ and $\mathbf{v}_2$; the
\emph{cosine similarity} is:
\begin{equation}
  \mathrm{CS}(\mathbf{v}_1,\mathbf{v}_2)
  = \frac{\mathbf{v}_1 \cdot \mathbf{v}_2}
         {\lVert \mathbf{v}_1 \rVert \,\lVert \mathbf{v}_2 \rVert } \, .
\end{equation}

We compute the \emph{cosine similarity} for all valid sentence--pair
combinations between the two segments and take the \emph{arithmetic mean} across all valid sentence pairs as the final similarity score for that sample.

\noindent\textbf{Accuracy rate (LLMJudge)} In computing the accuracy rate, we adopt LLMJudge with few-shot prompting, generating an explainable binary decision for each sample from which the accuracy rate is computed. Specifically, we design a unified LLMJudge instruction template that stipulates three criteria: first, whether the core idea is consistent with the reference; second, whether the key elements of the explanatory information can be located in and supported by the reference; and whether any fabricated content or information not covered by the reference is present. To stabilize decision boundaries, the template includes a small set of examples covering both correct and incorrect cases, and the judge model is invoked directly during evaluation to make the decision. The evaluation runs at the sample level: for each sample the judge returns a label, recorded as \(\mathrm{Pass}=1\) corresponding to [Correct] or  \(\mathrm{Pass}=0\) corresponding to [Incorrect]; we then aggregate within each task over samples to obtain the task level accuracy rate.
\begin{equation}
  \mathrm{Accuracy}
  = \frac{\text{number of Pass samples}}{\text{total number of samples}} \, .
\end{equation}

%% file: Section/experiment.tex
\section{Experiment}
\subsection{\textbf{Setup}}
  To systematically evaluate models’ retrieval, generation, and compliance judgment abilities in policy contexts, we select a representative suite that spans closed source and open source models as well as reasoning and chat paradigms. We then compare these models under a unified evaluation pipeline with shared hyperparameters. Selection follows the same basic criteria: recency, meaning the latest or currently available versions; diversity, meaning both reasoning oriented and general chat architectures; and practical scale, meaning a focus on small and medium models because of compute limits. The full list is:

  \textbf{Closed-source:}  \textbf{Reasoning} Gemini-2.5-Pro-Preview-06-05 \citep{gemini15}, Claude-3.7-Sonnet-Thinking \citep{claude3card}, o4-mini \citep{openai_o3o4_card}, o3-2025-04-16 \citep{openai_o3o4_card}.\\
\textbf{Chat} Claude-3.7-Sonnet-Latest \citep{claude3card}, GPT-4.1-20250414 \citep{gpt4tr}.

\textbf{Open-source}
 \textbf{Reasoning} DeepSeek-R1 \citep{deepseekr1}, DeepSeek-R1-Distill-Llama-8B \citep{deepseekr1distill}, Qwen3-8B \citep{qwen3tr}.\\
\textbf{Chat}  DeepSeek-V3 \citep{deepseekv3}, Llama3.1-8B-Instruct \citep{llama3paper,llama31card}.

  These models cover mainstream approaches in both ecosystems and form contrasted baselines along the reasoning and dialogue axes. Detailed experimental settings are provided in the appendix \ref{app1}.

\subsection{\textbf{Evaluation Result and Analysis}}

\input{Section/table_model}
    \textbf{Some open-source models exhibit discrepancies between literal semantic alignment and genuine task intent comprehension.}   It is noteworthy that high semantic similarity does not necessarily indicate the model's true understanding of task intent. Table \ref{table_model} show that Qwen3-8B achieved a semantic similarity of 0.64 but only 0.53 accuracy; while LLaMA3.1-8B-Instruct achieved an even higher semantic similarity of 0.71 yet scored only 0.40 in accuracy. This indicates that while some open-source models can obtain high similarity scores by directly retrieving or parroting policy clauses, they fall short in actual task comprehension and reasoning. 
    This results in responses that appear reasonable on the surface but deviate substantially from the intended meaning. This reflects that some open-source models still lack deep semantic understanding and dynamic generation capabilities in scenario-based policy tasks.

    \textbf{The reasoning model demonstrated significantly superior overall accuracy.}  \Cref{fig:3} shows that reasoning model's overall performance is above average. More specifically, as shown in \Cref{table_model}, the open-source reasoning model DeepSeek-R1 achieved an overall accuracy rate of 0.77, whereas the chat model DeepSeek-V3 recorded only 0.53. More notably, within the same closed-source model series, the reasoning model Claude-3.7-Sonnet-Thinking achieved an overall accuracy rate of 0.68, markedly surpassing the chat model Sonnet-latest's 0.55. This indicates that government tasks heavily rely on the complete logical chain from comprehension and reasoning, and reasoning models can more consistently maintain high accuracy across tasks.

\input{Section/table_language}
    \textbf{Most Models Generally Perform Better in Handling English Policies than Chinese Policies.} The results in Table \ref{table_Lan} show that most models perform better in handling English policies. For example, o4-mini has an accuracy rate of 0.69 for English policies, while it is only 0.58 for Chinese, which is a significant difference. Similarly, Qwen3-8B has a weaker performance with an accuracy rate of 0.57 for English policies and only 0.48 for Chinese policies. Most of the models perform more consistently in processing tasks in English policies, indicating that there is still room for optimisation of the \textit{State-Of-The-Art} model at the multi-language level, and in particular, the improvement of Chinese task adaptation is crucial to model performance.

    \textbf{DeepSeek-R1 performs well in dealing with Chinese policies.}  As can be seen from the results in Table \ref{table_task_cn} in \ref{task_table}, DeepSeek-R1 achieves the highest accuracy rate of 0.79 in dealing with Chinese policies, showing relatively strong Chinese optimisation capabilities. In the three tasks of processing Chinese policies, as shown in Table \ref{table_task_cn} in \ref{task_table}, DeepSeek-R1 achieves accuracy rates of 0.75, 0.84 and 0.77 respectively, which are significantly higher than the performance of other models under Chinese policies, showing its stability and accuracy in Chinese policy processing tasks. This suggests that multicorpus adaptation and multilingual optimisation may be able to improve the model's multilingual adaptability and robustness in multilingual contexts.

    \input{Figure/leida}
    \textbf{Reasoning models have better robustness across tasks.} As demonstrated in \Cref{fig:3}, reasoning models exhibit considerable advantages in cross-task stability. For instance, in the three Chinese tasks, the reasoning model average accuracy rates are 0.60, 0.64, and 0.61, respectively, showing only minor performance fluctuations, which indicates excellent stability. In contrast, the chat model average accuracy rates in the same three types of tasks are 0.56, 0.45, and 0.49, with a larger fluctuation in accuracy, and a noticeable performance decrease in the ``Solution Generation'' and ``Compliance Judgment'' tasks. This shows that reasoning models are more robust in handling complex tasks and ensuring accuracy and reliability in tasks that require understanding policy details, making inferences and identifying compliance boundaries.
    

   \textbf{Compliance judgment Class Tasks are More Challenging.} According to the experimental result data in \Cref{fig:3}, in terms of performance on different task types, the compliance judgment class tasks are significantly more difficult. As demonstrated in \Cref{fig:3}, the compliance judgment tasks exhibited the lowest overall mean accuracy rates in both the Chinese and English policy scenarios, at 0.56 and 0.60 respectively. Specifically, as shown in Table \ref{table_task_en} in \ref{task_table}, Claude-3.7-Sonnet-Thinking's accuracy rate of 0.56 on the compliance judgment class task is significantly lower than the accuracy rate of 0.74 obtained on the clause retrieval task and the accuracy rate of 0.67 obtained on the scenario generation task in English policy scenario. Accordingly, we believe that when the problem involves clause boundary conditions, the model needs to process more contextual information and perform complex reasoning chains, which requires stronger reasoning and comprehension abilities, and makes it much more difficult for models to deal with these tasks.

 \begin{figure}[t!] 
    \centering 
    \includegraphics[width=1\textwidth]{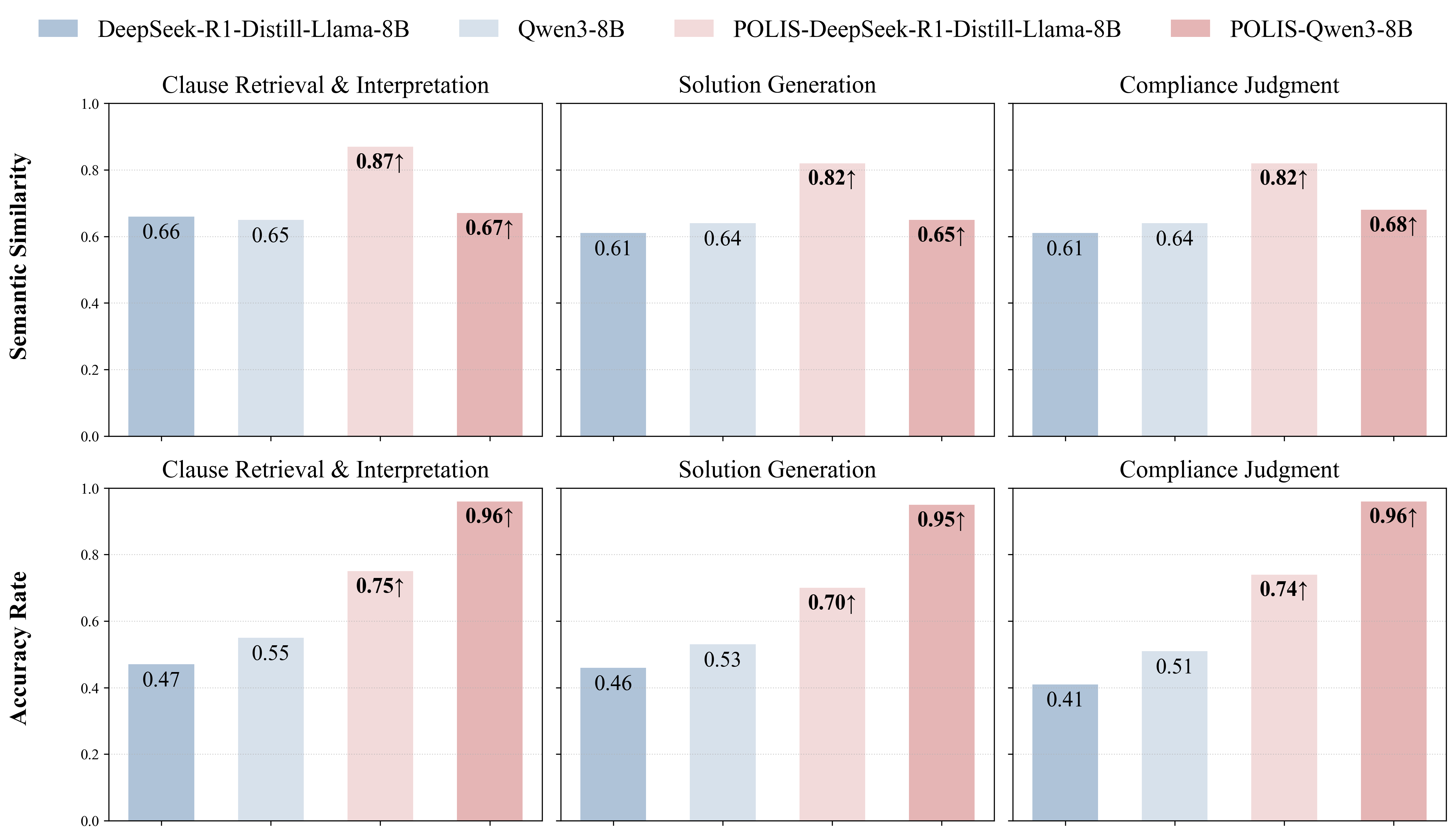} 
    \caption{\textbf{Task-oriented performance of base and \textsc{POLIS-tuned} models.}
    Bar plots compare four models (DeepSeek-R1-Distill-Llama-8B, Qwen3-8B, \textbf\textsc{POLIS}-DeepSeek-R1-Distill-Llama-8B, \textbf\textsc{POLIS}-Qwen3-8B) on three tasks: \emph{Clause Retrieval \& Interpretation}, \emph{Solution Generation}, and \emph{Compliance Judgment}.
    The top row reports \textit{Semantic Similarity}, and the bottom row reports \textit{Accuracy Rate}; higher values indicate better performance.
    Numbers atop bars are the corresponding scores, obtained under the unified evaluation pipeline.}
    \label{finetuned_result} 
    \end{figure}
    
\subsection{\textbf{Training Result and Analysis}}  
    \textbf{Fine--tuning} To evaluate the effectiveness of \textsc{POLIS--Bench}, we conduct LoRA fine--tuning on two open-source models: Qwen3-8B and DeepSeek-R1-distill-llama-8b that performed well in our preliminary evaluation. These models were trained on the training set of \textsc{POLIS--Bench} dataset and subsequently evaluated on the test set. As shown in Figure \ref{finetuned_result}, \textsc{POLIS}-Qwen3-8b, which was fine-tuned on Qwen3-8b, exhibited a significant improvement in accuracy rate, achieving 0.96,0.95 and 0.96 respectively in 3 tasks, while maintaining semantic similarity with the mean score of 0.67. Compared with the base model Qwen3-8B, which gets accuracy rates of 0.55,o.53 and 0.51 in 3 tasks and mean semantic similarity score of 0.64, \textsc{POLIS}-Qwen3-8b achieved an average improvement of 80\% in accuracy, accompanied by a slight improvement in semantic coverage. Meanwhile, \textsc{POLIS}-DeepSeek-R1-distill-llama-8b, which fine-tuned on DeepSeek-R1-distill-llama-8b, shows a notable increase in semantic similarity, achieving an average score of 0.84 and its accuracy score has been elevated by 0.28 above its backbone model, which represents a 62\% increase. Both \textsc{POLIS} models achieve remarkable evaluation outcomes that remain competitive with several strong general-purpose baselines, and even meet or exceed those of several leading closed-source general models, such as the GPT series, on both the Chinese and English subsets. This suggests that task-aligned fine-tuning on the POLIS-Bench dataset consistently improves multi-task performance within policy-oriented settings.
    
    \textbf{General capability evaluation} To verify that task-aligned fine-tuning for government tasks does not compromise a model’s general capabilities, we conducted a controlled evaluation on the \textit{GPQA-Diamond} benchmark\cite{rein2024gpqa}. \textit{GPQA-Diamond} is a high-difficulty, multi-disciplinary multiple-choice set spanning biology, physics, and chemistry, with four options per question and a 25\% random-guess baseline. In the \textit{GPQA} series, \textit{GPQA-Diamond} is particularly stringent that PhD-level experts can only achieve about 65\% accuracy, while skilled non-experts with web access reach about 34\%. That makes it effective for testing advanced reasoning and scientific general knowledge. We used the official Diamond subset (198 questions) for closed-book testing with consistent 5-shot prompt and fixed inference hyperparameters.
    
    \input{Section/GPQA-table}
    As shown in the Table \ref{GPQA-Diamond}, Qwen3-8B maintains the same Diamond accuracy before and after fine-tuning (36.87\%), DeepSeek-distill-Llama3-8B improves from 28.3\% to 30.4\% after fine-tuning. That indicates that task-aligned fine-tuning does not weaken its reasoning and knowledge-retrieval ability on cross-disciplinary, high-difficulty multiple choice. Given the difficulty and ceiling of \textit{GPQA-Diamond}, these differences lie within a reasonable range of statistical fluctuation. The results indicate that the fine-tuned models preserve downstream task performance and that their core capabilities—such as advanced reasoning—remain intact, supporting robustness and broad applicability in real-world deployments.

 \begin{figure}[h!] 
    \centering 
    \includegraphics[width=0.85\textwidth]{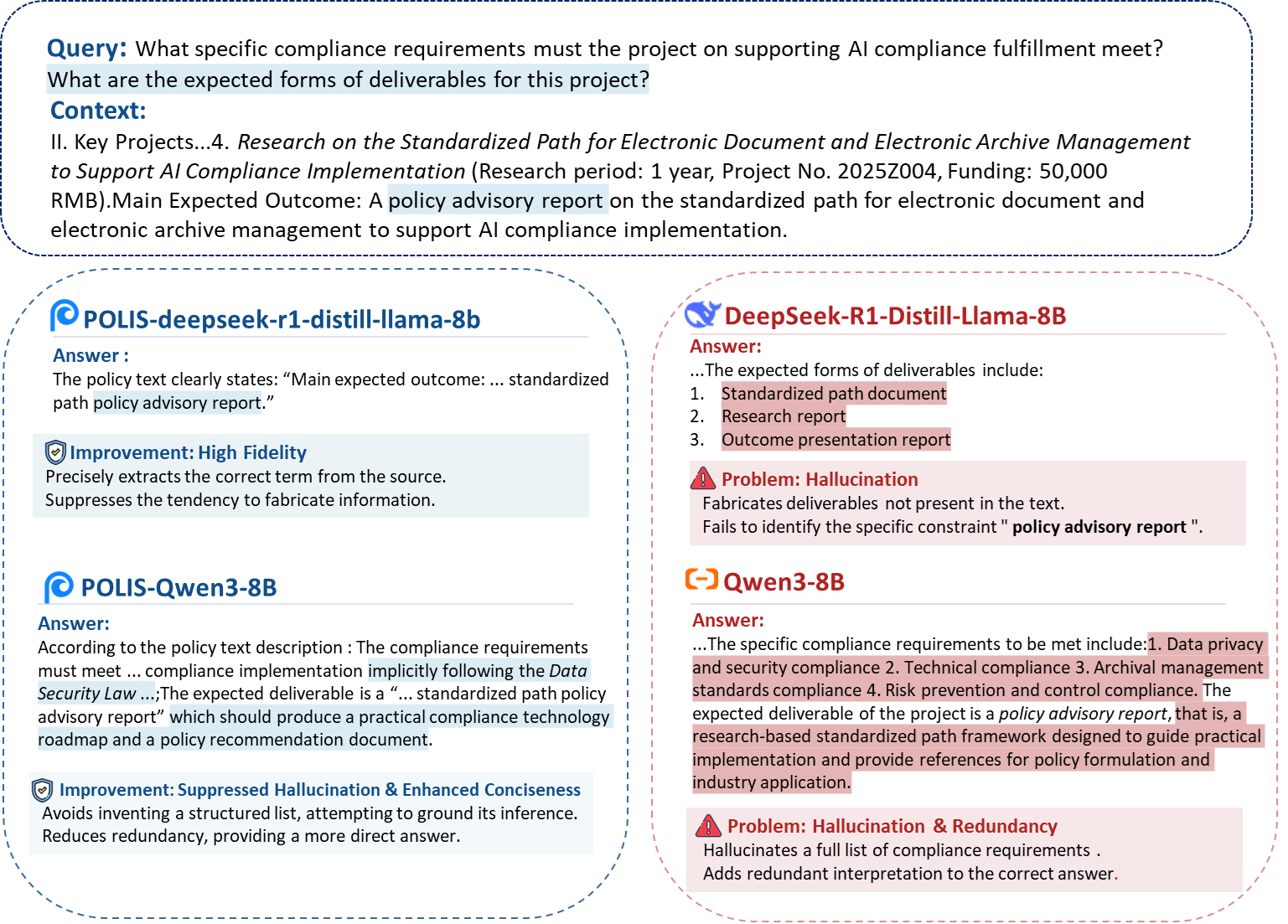} 
    \caption{\textbf{Case Study on Compliance Judgment Task}}
    \label{case study 1} 
    \end{figure}

 \begin{figure}[h!] 
    \centering 
    \includegraphics[width=0.85\textwidth]{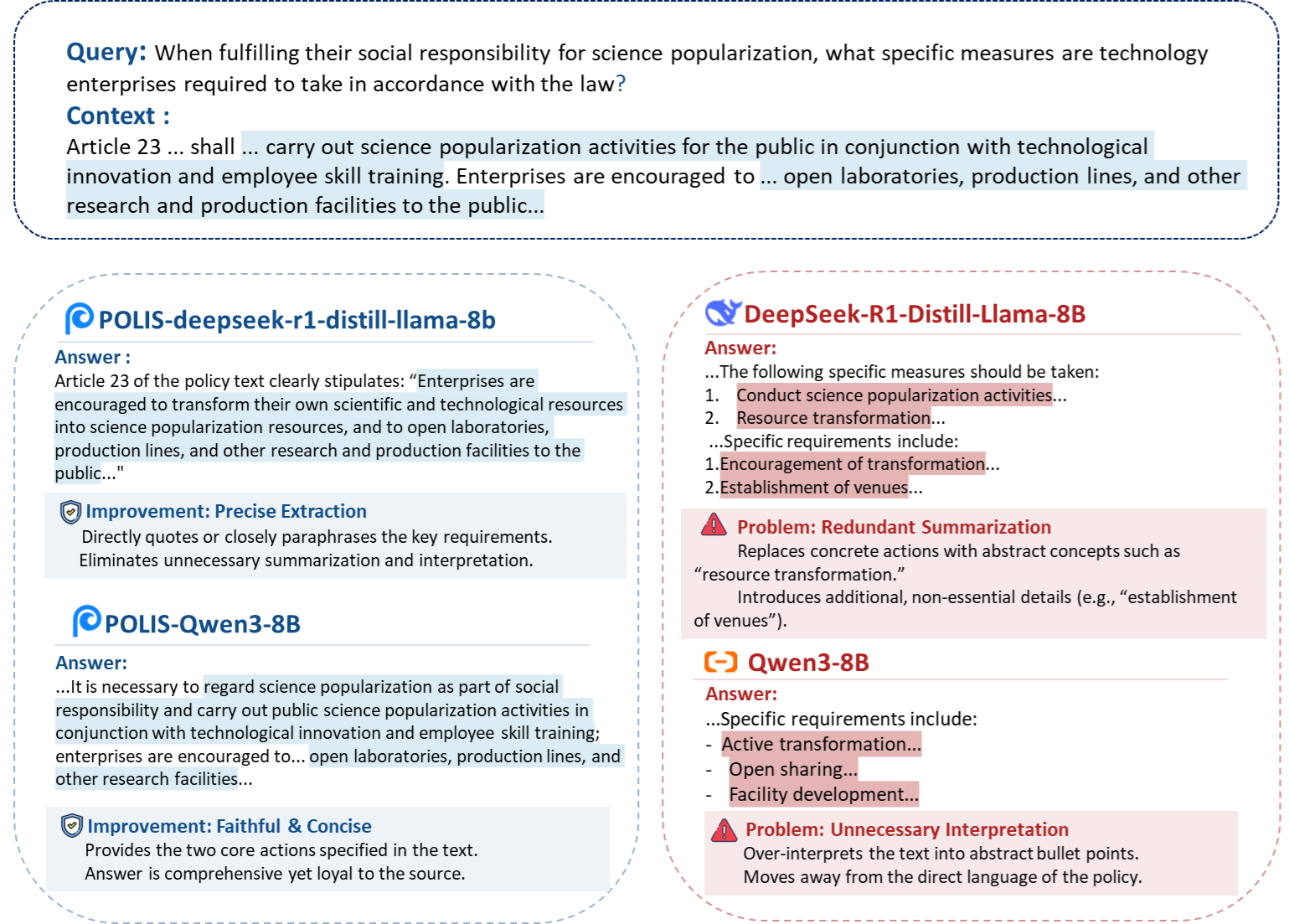} 
    \caption{Case Study on Solution Generation Task}
    \label{case study 2} 
    \end{figure}
    
\subsection{\textbf{Case Study}}
To more intuitively illustrate the performance improvements achieved through task-aligned parameter-efficient LoRA fine-tuning based on \textsc{\textbf{POLIS-Bench}}, we conducted a case study comparing the outputs of the fine-tuned model with those of two baseline models: Qwen3-8B and DeepSeek-R1-Distill-Llama-8B. This comparative analysis was designed to demonstrate the fine-tuned model’s enhanced capacity to accurately capture task-specific requirements in policy-related contexts.
In the first case as shown in \Cref{case study 1}, the task required the model to accurately delineate the boundaries of missing information as specified by the policy text. However, because the text did not provide a detailed elaboration of the “compliance requirements”,   both baseline models were adversely influenced by contextual ambiguity and incorrectly fabricated non-existent clauses such as “data privacy compliance.” This behavior exemplifies a typical hallucination phenomenon, leading to the generation of factually incorrect content.
In the second case as shown in \Cref{case study 2}, both baseline models introduced redundant interpretations and generalized summaries in their responses. Although the core information was preserved, the inclusion of such extraneous content undermined the conciseness and faithfulness of their outputs relative to the source text.

Collectively, these results indicate that task-aligned LoRA fine-tuning grounded in \textsc{POLIS--Bench}, when implemented under a unified instruction template, effectively mitigates persistent issues such as factual hallucination and redundant generation observed in base models. In key tasks---including Compliance Judgment and Solution Generation---the fine-tuned model demonstrates superior textual fidelity and task alignment. These findings provide empirical evidence that the proposed approach can achieve performance comparable to, or exceeding, that of strong proprietary baselines while maintaining parameter efficiency, cost-effectiveness, and enhanced deployability.

%% file: Section/table_model.tex
\begin{wraptable}{r}{8cm}
\vspace*{-\baselineskip}
\centering
\caption{\textbf{Overall Evaluation Results.} Quantitative results on \textbf{\textsc{POLIS-Bench}} across closed-source \& open-source and reasoning \& chat model families. 
For each language, the table reports two metrics: \texttt{semantic similarity} (mean semantic similarity) and \texttt{accuracy rate} (task completion accuracy measured by LLMJudge) under a unified evaluation pipeline.
Cells highlighted in \colorbox{bestgreen}{red} indicate the \textbf{highest score} within each column, and \colorbox{secondgreen}{blue} indicates the \textbf{second highest}.}

\label{table_model}
\resizebox{0.5\textwidth}{!}{%
\begin{tabular}{lcc}
\hline
\multicolumn{1}{c}{\textbf{Model}} & \textbf{similarity\_mean} & \textbf{accuracy\_rate}      \\ \hline
\multicolumn{3}{c}{\cellcolor[HTML]{FFFFFF}\textbf{Closed-source   Reasoning Model}}          \\
Gemini-2.5-Pro-Preview-06-05       & 0.68                      & \cellcolor[HTML]{AFC3D8}0.71 \\
Claude-3.7-Sonnet-Thinking   & 0.65                         & 0.68                         \\
o4-mini                      & 0.66                         & 0.63                         \\
o3-2025-04-16                & 0.65                         & 0.70                         \\
\multicolumn{3}{c}{\cellcolor[HTML]{FFFFFF}\textbf{Closed-source Chat   Model}}               \\
Claude-3.7-Sonnet-Latest     & 0.69                         & 0.55                         \\
GPT-4.1-20250414             & \cellcolor[HTML]{AFC3D8}0.70 & 0.68                         \\ \hline
\multicolumn{3}{c}{\cellcolor[HTML]{FFFFFF}\textbf{Open-source   Reasoning Model}}            \\
DeepSeek-R1                  & 0.65                         & \cellcolor[HTML]{E5B5B5}0.77 \\
DeepSeek-R1-Distill-Llama-8B & 0.63                         & 0.45                         \\
Qwen3-8B                     & 0.64                         & 0.53                         \\
\multicolumn{3}{c}{\cellcolor[HTML]{FFFFFF}\textbf{Open-source Chat   Model}}                 \\
DeepSeek-V3                  & 0.67                         & 0.52                         \\
Llama3.1-8B-Instruct         & \cellcolor[HTML]{E5B5B5}0.71 & 0.40                         \\ \hline
\end{tabular}%
}
\end{wraptable}

%% file: Section/table_language.tex
\begin{table}[h!]
\centering
\caption{\textbf{Cross-language Evaluation Results.} Quantitative results on \textbf{\textsc{POLIS-Bench}} under \textbf{Chinese (CN)} and \textbf{English (EN)} policy contexts, covering closed-source \& open-source and reasoning \& chat model families.
For each language, the table reports two metrics: \texttt{semantic similarity} (mean semantic similarity) and \texttt{accuracy\_rate} (task completion accuracy measured by LLMJudge) under a unified evaluation pipeline.
Cells highlighted in \colorbox{bestgreen}{red} indicate the \textbf{highest score} within each subcolumn, and \colorbox{secondgreen}{blue} indicates the \textbf{second highest}.}

\label{table_Lan}
\resizebox{0.7\textwidth}{!}{%
\begin{tabular}{lcccc}
\hline
\multicolumn{1}{c}{} & \multicolumn{2}{c}{\textbf{CN}} & \multicolumn{2}{c}{\textbf{EN}} \\
\multicolumn{1}{c}{\multirow{-2}{*}{\textbf{Model}}} & \textbf{semantic similarity} & \textbf{accuracy\_rate} & \textbf{semantic similarity} & \textbf{accurate\_rate} \\ \hline
\multicolumn{5}{c}{\textbf{Closed-Source   Reasoning Model}} \\
Gemini-2.5-Pro-Preview-06-05 & 0.63 & \cellcolor[HTML]{AFC3D8}0.72 & \cellcolor[HTML]{E5B5B5}0.74 & 0.70 \\
Claude-3.7-Sonnet-Thinking & 0.60 & 0.68 & 0.71 & 0.68 \\
o4-mini & 0.63 & 0.58 & 0.69 & 0.69 \\
o3-2025-04-16 & 0.60 & 0.67 & 0.71 & \cellcolor[HTML]{AFC3D8}0.74 \\
\multicolumn{5}{c}{\textbf{Closed-Source Chat   Model}} \\
Claude-3.7-Sonnet-Latest & \cellcolor[HTML]{E5B5B5}0.71 & 0.53 & 0.68 & 0.57 \\
GPT-4.1-20250414 & 0.66 & 0.62 & \cellcolor[HTML]{E5B5B5}0.74 & \cellcolor[HTML]{E5B5B5}0.75 \\ \hline
\multicolumn{5}{c}{\textbf{Open-Source   Reasoning Model}} \\
DeepSeek-R1 & 0.59 & \cellcolor[HTML]{E5B5B5}0.79 & \cellcolor[HTML]{AFC3D8}0.72 & \cellcolor[HTML]{E5B5B5}0.75 \\
DeepSeek-R1-Distill-Llama-8B & 0.58 & 0.39 & 0.69 & 0.52 \\
Qwen3-8B & 0.58 & 0.48 & 0.71 & 0.59 \\
\multicolumn{5}{c}{\textbf{Open-Source Chat   Model}} \\
DeepSeek-V3 & 0.65 & 0.49 & 0.71 & 0.56 \\
Llama3.1-8B-Instruct & \cellcolor[HTML]{AFC3D8}0.69 & 0.33 & \cellcolor[HTML]{E5B5B5}0.74 & 0.48 \\ \hline
\end{tabular}%
}
\end{table}

%% file: Figure/leida.tex
\begin{wrapfigure}{r}{8cm}
\centering
\includegraphics[width=0.5\textwidth]{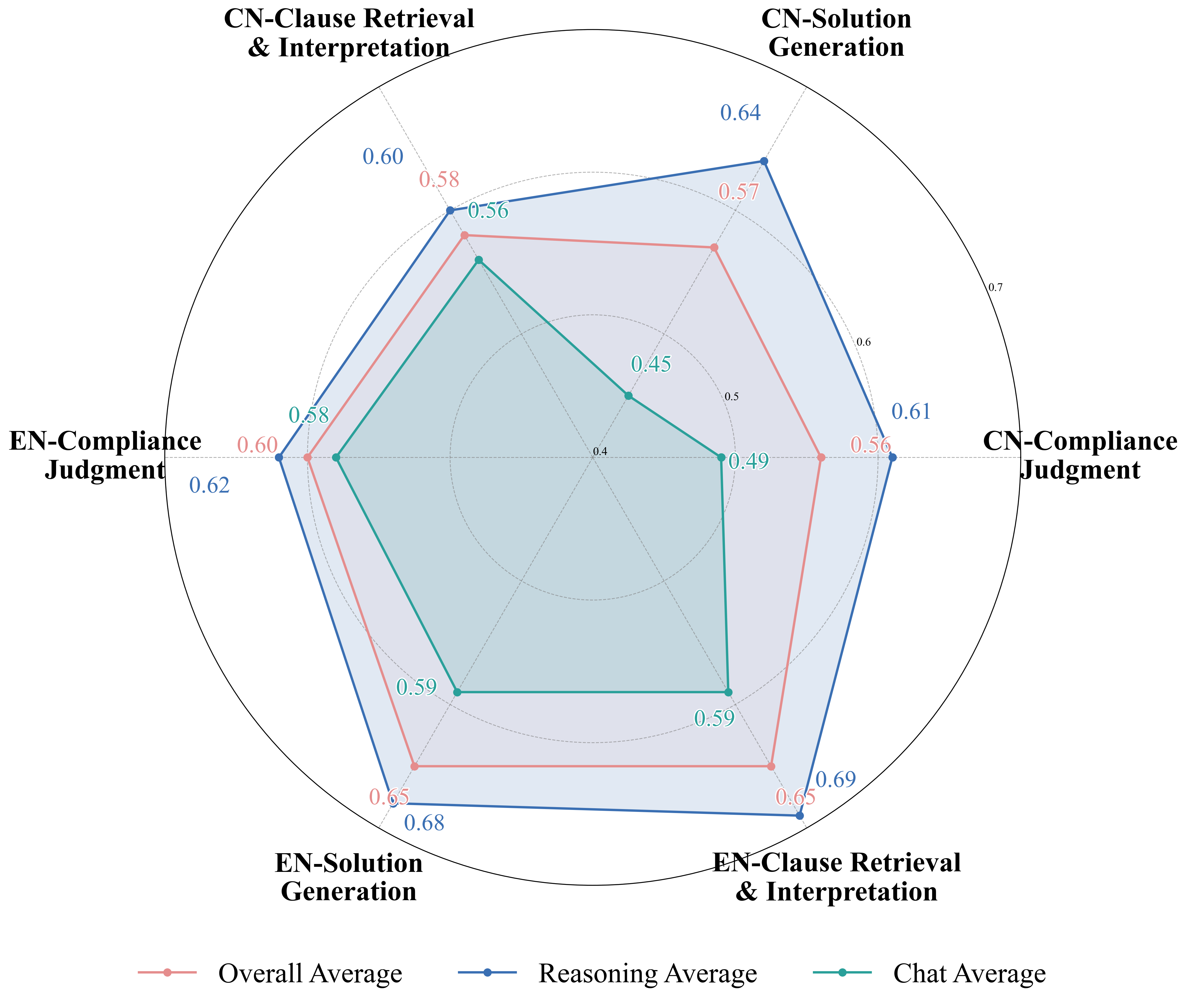}
\caption{Accuracy Rate Distribution. This chart illustrates the accuracy rate distribution for the average performance of Reasoning Models (Reasoning Average), Chat Models (Chat Average), and the overall mean (Overall Average) across the three bilingual policy tasks on \textsc{POLIS-Bench}.}
\label{fig:3}
\end{wrapfigure}

%% file: Section/GPQA-table.tex

\begin{wraptable}{r}{8cm}
\centering
\caption{The general capability evaluation result of POLIS-DeepSeek-R1-Distill-Llama-8B, POLIS-Qwen3-8B and their backbone models on \textit{GPQA-Diamond} benchmark}
\label{GPQA-Diamond}
\resizebox{0.5\textwidth}{!}{%
\begin{tabular}{lc}
\hline
\multicolumn{1}{c}{\textbf{Model}}          & \textbf{Accuracy} \\ \hline
DeepSeek-R1-Distill-Llama-8B                & 0.28              \\
Qwen3-8B                                    & 0.37              \\
\textbf{POLIS-DeepSeek-R1-Distill-Llama-8B} & 0.3               \\
\textbf{POLIS-Qwen3-8B}                     & 0.37              \\ \hline
\end{tabular}%
}
\end{wraptable}

%% file: Section/conclusion.tex
\section{\textbf{Conclusion}}
In this work, we introduced \textsc{POLIS-Bench}, a systematic benchmark for policy-oriented generation that couples an up-to-date bilingual policy corpus with three scenario-grounded tasks：Clause Retrieval and Interpretation, Solution Generation, and Compliance Judgment. And a unified dual-metrics evaluation pipeline for semantic proximity and task correctness. Across a representative suite of closed-- and open--source models, our large scale study reveals a consistent performance hierarchy: reasoning models deliver stronger and more stable results than chat models, with compliance judgment posing the greatest challenge among the three task families. Building on the benchmark, task-aligned fine-tuning leads to clear multi-task gains. In particular, our lightweight adapted model attains parity with or exceeds strong closed-source baselines at lower cost, while preserving general reasoning ability as verified on GPQA-Diamond. 

Looking forward, we see two priorities. First, broaden coverage. Extend to additional jurisdictions, policy domains, and multilingual settings, and stress-test models on harder, boundary-heavy cases to better characterize retrieval, generation, and compliance trade-offs. Second, institute dynamic updates and governance: continuously refresh the corpus with newly released regulations, calibrate LLMJudge prompts against expert rubrics, and strengthen contamination auditing to safeguard integrity and reproducibility. With wider scope and stronger dataset governance, future iterations can provide a more comprehensive and durable standard for evaluating and improving policy-aware language intelligence.

%% file: Section/appendix.tex
\newpage
\section{EXPERIMENTAL SETUP}
\label{app1}
  We adopt low-stochasticity, reproducible settings in the unified evaluation pipeline. In both the answer-generation stage and the judge--decision stage, we use identical inference hyperparameters: single-pass inference (no multi-sample averaging or voting), low-temperature sampling (temperature = 0.1), disabled log-probability return (logprobs = False), and max\_tokens set to each model’s maximum permitted output length. To reduce noise caused by interface fluctuations, both stages enable up to five failure retries (triggered upon exceptions or unparseable responses, stopping immediately upon success). Inputs strictly follow length constraints: over--long policy originals are already standardized and segmented during data construction, ensuring that the per--sample context does not exceed the processable window, thereby avoiding output truncation and preserving answer completeness and validity.
  
  In the LLMJudge step, we use QwQ-32B \citep{qwq32b_blog,qwq32b_hf} as the judge model and apply a fixed few-shot template; the judge’s output is hard-constrained to the binary labels "[Correct]/[Incorrect]". All samples and all three tasks (Clause Retrieval \& Interpretation, Solution Generation, Compliance Judgment) are evaluated under the same prompts and hyperparameters, ensuring that scores across tasks and models are comparable and reproducible. These settings minimize the impact of sampling stochasticity without sacrificing the strictness of judgments.

\section{MORE EXPERIMENTAL RESULTS AND DISCUSSIONS}
\subsection{Task-level Performance Distribution under Bilingual Policies}
    \label{task_table}
    The following tables, Table 3 and Table 4, provide a detailed, task-wise breakdown of the models' performance on the POLIS-Bench test set, disaggregated by language (Chinese and English). For each task—Clause Retrieval \& Interpretation, Solution Generation, and Compliance Judgment—the quantitative results report both the mean semantic similarity and the LLM.Judge accuracy rate. These data further illustrate the cross-task and cross-language stability and fluctuations observed across the closed-source (Reasoning and Chat) and open-source (Reasoning and Chat) model families.
\input{Section/table_task_cn}
\input{Section/table_task_en}


%% file: Section/table_task_cn.tex
\begin{table}[h!]
\centering
\caption{\textbf{Task-wise Results under Chinese Policies.} Quantitative results on \textbf{\textsc{POLIS-Bench}} (CN) covering closed-source \& open-source and reasoning \& chat model families across three tasks. 
For each task, the table reports \texttt{semantic similarity} and \texttt{accuracy\_rate} under a unified evaluation pipeline. 
Cells highlighted in \colorbox{bestgreen}{red} indicate the \textbf{highest} value within each metric column, and \colorbox{secondgreen}{blue} indicates the \textbf{second highest}.}
\label{table_task_cn}
\resizebox{\textwidth}{!}{%
\begin{tabular}{lcccccc}
\hline
\multicolumn{1}{c}{} & \multicolumn{2}{c}{\textbf{CN-Clause Retrieval   \& Interpretation}} & \multicolumn{2}{c}{\textbf{CN-solution generation}} & \multicolumn{2}{c}{\textbf{CN-compliance judgment}} \\
\multicolumn{1}{c}{\multirow{-2}{*}{\textbf{Model}}} & semantic similarity & accurate rate & semantic similarity & accurate rate & semantic similarity & accurate rate \\ \hline
\multicolumn{7}{c}{\cellcolor[HTML]{FFFFFF}\textbf{Closed-source  Reasoning  Model}} \\
Gemini-2.5-Pro-Preview-06-05 & 0.64 & \cellcolor[HTML]{E5B5B5}0.75 & 0.62 & 0.68 & 0.62 & \cellcolor[HTML]{AFC3D8}0.72 \\
Claude-3.7-Sonnet-Thinking & 0.64 & 0.65 & 0.58 & \cellcolor[HTML]{AFC3D8}0.73 & 0.59 & 0.65 \\
o4-mini & 0.66 & 0.54 & 0.60 & 0.63 & 0.61 & 0.59 \\
o3-2025-04-16 & 0.63 & 0.66 & 0.58 & 0.70 & 0.59 & 0.64 \\
\multicolumn{7}{c}{\textbf{Closed-source  Chat  Model}} \\
Claude-3.7-Sonnet-Latest & \cellcolor[HTML]{E5B5B5}0.71 & 0.63 & \cellcolor[HTML]{E5B5B5}0.71 & 0.45 & \cellcolor[HTML]{E5B5B5}0.69 & 0.59 \\
GPT-4.1-20250414 & \cellcolor[HTML]{AFC3D8}0.69 & \cellcolor[HTML]{AFC3D8}0.68 & 0.63 & 0.58 & 0.65 & 0.59 \\ \hline
\multicolumn{7}{c}{\textbf{Open-source  Reasoning  Model}} \\
DeepSeek-R1 & 0.61 & \cellcolor[HTML]{E5B5B5}0.75 & 0.57 & \cellcolor[HTML]{E5B5B5}0.84 & 0.59 & \cellcolor[HTML]{E5B5B5}0.77 \\
DeepSeek-R1-Distill-Llama-8B & 0.59 & 0.38 & 0.56 & 0.40 & 0.58 & 0.38 \\
Qwen3-8B & 0.57 & 0.46 & 0.59 & 0.49 & 0.59 & 0.49 \\
\multicolumn{7}{c}{\textbf{Open-source  Chat  Model}} \\
Llama3.1-8B-Instruct & \cellcolor[HTML]{E5B5B5}0.71 & 0.41 & \cellcolor[HTML]{AFC3D8}0.67 & 0.28 & \cellcolor[HTML]{AFC3D8}0.67 & 0.29 \\
DeepSeek-V3 & 0.68 & 0.52 & 0.62 & 0.48 & 0.65 & 0.48 \\ \hline

\end{tabular}%
}
\end{table}

%% file: Section/table_task_en.tex
\vspace*{-50\baselineskip}
\begin{table}[t!]
\centering
\caption{\textbf{Task-wise Results under English Policies.} Quantitative results on \textbf{\textsc{POLIS-Bench}} (EN) covering closed-source \& open-source and reasoning \& chat model families across three tasks. 
For each task, the table reports \texttt{semantic similarity} and \texttt{accuracy\_rate} under a unified evaluation pipeline. 
Cells highlighted in \colorbox{bestgreen}{red} indicate the \textbf{highest} value within each metric column, and \colorbox{secondgreen}{blue} indicates the \textbf{second highest}.}

\label{table_task_en}
\resizebox{\textwidth}{!}{%
\begin{tabular}{lcccccc}
\hline
\multicolumn{1}{c}{} & \multicolumn{2}{c}{\textbf{EN-Clause Retrieval   \& Interpretation}} & \multicolumn{2}{c}{\textbf{EN-solution generation}} & \multicolumn{2}{c}{\textbf{EN-compliance judgment}} \\
\multicolumn{1}{c}{\multirow{-2}{*}{\textbf{Model}}} & semantic similarity & accurate rate & semantic similarity & accurate rate & semantic similarity & accurate rate \\ \hline
\multicolumn{7}{c}{\cellcolor[HTML]{FFFFFF}\textbf{Closed-source  Reasoning  Model}} \\
Gemini-2.5-Pro-Preview-06-05 & \cellcolor[HTML]{AFC3D8}0.78 & \cellcolor[HTML]{E5B5B5}0.69 & \cellcolor[HTML]{E5B5B5}0.71 & 0.70 & \cellcolor[HTML]{E5B5B5}0.73 & \cellcolor[HTML]{AFC3D8}0.70 \\
Claude-3.7-Sonnet-Thinking & 0.75 & \cellcolor[HTML]{AFC3D8}0.74 & 0.67 & \cellcolor[HTML]{AFC3D8}0.75 & \cellcolor[HTML]{AFC3D8}0.70 & 0.56 \\
o4-mini & 0.73 & 0.72 & 0.65 & 0.74 & 0.68 & 0.61 \\
o3-2025-04-16 & 0.75 & \cellcolor[HTML]{E5B5B5}0.75 & 0.68 & \cellcolor[HTML]{AFC3D8}0.75 & \cellcolor[HTML]{AFC3D8}0.70 & 0.69 \\
\multicolumn{7}{c}{\textbf{Closed-source  Chat  Model}} \\
Claude-3.7-Sonnet-Latest & \cellcolor[HTML]{E5B5B5}0.73 & 0.61 & \cellcolor[HTML]{E5B5B5}0.62 & 0.53 & \cellcolor[HTML]{E5B5B5}0.64 & 0.57 \\
GPT-4.1-20250414 & \cellcolor[HTML]{AFC3D8}0.77 & \cellcolor[HTML]{E5B5B5}0.75 & \cellcolor[HTML]{E5B5B5}0.71 & \cellcolor[HTML]{E5B5B5}0.76 & \cellcolor[HTML]{E5B5B5}0.73 & \cellcolor[HTML]{AFC3D8}0.72 \\ \hline
\multicolumn{7}{c}{\textbf{Open-source  Reasoning  Model}} \\
DeepSeek-R1 & 0.75 & \cellcolor[HTML]{E5B5B5}0.72 & \cellcolor[HTML]{AFC3D8}0.70 & \cellcolor[HTML]{E5B5B5}0.76 & \cellcolor[HTML]{AFC3D8}0.70 & \cellcolor[HTML]{E5B5B5}0.76 \\
DeepSeek-R1-Distill-Llama-8B & 0.73 & 0.55 & 0.66 & 0.51 & 0.68 & 0.47 \\
Qwen3-8B & 0.74 & 0.63 & 0.69 & 0.57 & \cellcolor[HTML]{AFC3D8}0.70 & 0.54 \\
\multicolumn{7}{c}{\textbf{Open-source  Chat  Model}} \\
Llama3.1-8B-Instruct & \cellcolor[HTML]{E5B5B5}0.79 & 0.50 & \cellcolor[HTML]{E5B5B5}0.71 & 0.45 & \cellcolor[HTML]{E5B5B5}0.73 & 0.48 \\
DeepSeek-V3 & \cellcolor[HTML]{AFC3D8}0.78 & 0.51 & 0.68 & 0.60 & \cellcolor[HTML]{AFC3D8}0.70 & 0.55 \\ \hline
\end{tabular}%
}
\end{table}